\pdfoutput=1
\documentclass[11pt]{article}

\usepackage[preprint]{acl}

\usepackage{times}
\usepackage{latexsym}

\usepackage[T1]{fontenc}
\usepackage[utf8]{inputenc}

\usepackage{microtype}

\usepackage{inconsolata}

\usepackage{graphicx}
\usepackage{enumitem} 

\usepackage{microtype}
\usepackage{makecell}
\usepackage{svg}
\usepackage{subcaption}
\usepackage{booktabs}
\usepackage{amsmath}
\usepackage{amssymb}
\usepackage{amsfonts}
\usepackage[super]{nth}
\usepackage{cancel}
\usepackage[normalem]{ulem}

\DeclareMathOperator{\wavto}{Wav2vec}
\DeclareMathOperator{\fnn}{FNN}
\DeclareMathOperator{\lstm}{LSTM}
\DeclareMathOperator{\bilstm}{bi-LSTM}
\DeclareMathOperator{\ctc}{CTC}

\title{When Can You Prune Your Network? \\ A Study of Intermediate Neurons in Multilingual Speech Parsing}

\author{Minnie Kabra, Benjamin Lecouteux, Maximin Coavoux \\
   Univ. Grenoble Alpes, CNRS, Grenoble INP, LIG, 38000 Grenoble, France\\
  \texttt{first.last@univ-grenoble-alpes.fr}
}

\begin{document}

\maketitle
\begin{abstract}

End-to-end speech parsing, a task recently proposed, consists in predicting both the transcription and the syntactic tree for a spoken utterance. 
Existing architectures for speech parsing often utilise intermediate neural networks. 
In this work, we examine the effectiveness of intermediate neural networks (NN) for parsing, and, specifically, what role do they play. We introduce a simpler end-to-end architecture for speech parsing, where we remove these intermediate NN units, reducing the parameters by 12\%,  while achieving comparable or better performance than prior method on both automatic speech recognition (ASR) and parsing. We demonstrate that intermediate NN units help reduce the \textit{representational gap} when the pre-trained encoder is frozen. We do a comprehensive evaluation of speech parsing on French, and medium-low resource languages Slovenian and Naija. We further investigate the impact of the training data size and intermediate layers of the pretrained speech encoder on speech parsing.

\end{abstract}

\section{Introduction}

End-to-end architectures for speech parsing \citep{pupier22_interspeech,kando24_interspeech} connect the pre-trained speech and the parsing models using intermediate neural network (NN) units like feedforward neural network (FNN), recurrent neural network (RNN), and long short-term memory (LSTM) units. These units increase the computational training time. However, the impact of these units has not been yet studied for parsing.

\begin{figure}
    \centering
    \resizebox{\columnwidth}{!}{
    \includegraphics{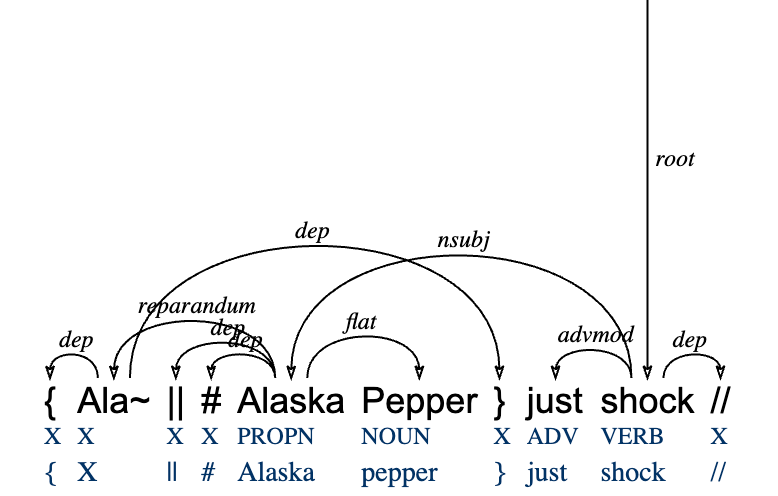}}
    \caption{Dependency parsing tree from the NaijaNSC treebank. \textsc{en}: Ala-... Alaska Pepper was shocked. The two lines under the wordform line are the sequence of POS and lemma tags respectively.}
    \label{fig:tree}
\end{figure}

Even though dependency parsing (Figure \ref{fig:tree}) has nowadays fewer applications in NLP, the design of accurate syntactic parsers is still relevant for applications to data-driven linguistics \citep{hull-dobrovoljc-2025-word,PeckBecker} whose analyses rely on syntactic annotations. However, treebanks of spontaneous speech, arguably the most naturalistic source of linguistic data, are scarcer than written treebanks.The parsing of spontaneous speech is currently not at a satisfactory level. Hence, high-quality speech parsers would open new perspectives for the study of syntax and for computational language documentation. More generally, besides these motivations, our work subscribes to the current necessity for the NLP community to focus more on speech as the most naturalistic source of linguistic data \citep{chrupala-2023-putting}.

Wav2tree \citep{pupier22_interspeech} is an end-to-end architecture for spontaneous speech parsing where the pre-trained speech model is fine-tuned jointly with the downstream parsing model, with intermediary NN units inserted between the two models. It takes speech as input, and produces both the transcription and the syntactic tree as the output. Due to this joint fine-tuning setup,  the representations learned by the speech encoder, as well as those produced by the intermediate NN units, is optimised for the downstream parsing model.

Our contributions are the following:
\begin{enumerate}
 \item We examine the role of intermediate neural network (NN) units in Wav2tree:
\begin{itemize}[leftmargin=*]
\item we introduce a new end-to-end architecture, \textit{Simplified Wav2tree} (Simplified W2T), for parsing speech to a dependency tree. It simplifies Wav2tree by removing these intermediate NN units, while generally matching or improving performance with greater efficiency.\footnote{\url{https://github.com/minniekabra/SpeechParsing} \label{link:speechparsing}} 
\item When the pre-trained model (encoder) is frozen within a joint parsing architecture, its representation remains optimised for the pre-training objective, and not for the downstream parsing task. We refer to this phenomenon as the \textit{representational gap} between the encoder and the parsing model. We demonstrate that intermediate neural network units help reduce this \textit{representational gap} between the frozen encoder and the parsing model.

\end{itemize} 

\item We carry out a comprehensive evaluation of dependency parsing of speech for the spoken treebanks of Naija, Slovenian, and French:
\begin{itemize}[leftmargin=*]
  \item We compare Simplified Wav2tree with baseline Wav2tree.
  \item We examine the performance of dependency parsing using representation from intermediate Wav2Vec2\cite{baevski2020wav2vec} layers as the input to both ASR and parsing model in the Simplified architecture. The performance of parsing remains stable or improves, increasing the parameter efficiency of the architecture.
  \item We do a comparison 
  with the pipeline approach 
  and gold transcriptions. Except for Slovenian, the end-to-end parsing performs better than the pipeline.
  
\end{itemize} 
\item  We analyze the impact of scaling training data on dependency parsing of speech. Through a two-stage experimental setup, we assess the amount of speech data required by the parsing model to attain a near-optimal performance with the Simplified architecture.

\item We present language-specific analyses for Naija and Slovenian. For Naija, we assess the usefulness of segmentation annotations for parsing. Our experiments show that they do not lead to improvements as they do when parsing transcriptions. For Slovenian, we release the code\textsuperscript{\ref{link:speechparsing}} to produce filtered \texttt{UD\_Slovenian-SST} v2.16 with a better speech-transcription alignment.
\end{enumerate}

\begin{table}[]
\centering
\resizebox{\columnwidth}{!}{%
\begin{tabular}{lllrr}
\toprule
\multicolumn{1}{c}{\textbf{Dataset}} &
  \multicolumn{1}{c}{\textbf{Section}} &
  \multicolumn{1}{c}{\textbf{Duration}} &
  \multicolumn{1}{c}{\textbf{Trees}} &
  \multicolumn{1}{c}{\textbf{Speakers}} \\
\midrule
             & Train & 6.78h & 7,268 & 67  \\
Naija        & Dev   & 0.87h & 990  & 10  \\
             & Test  & 0.91h & 972  & 11  \\
\midrule
             & Train & 4.88h & 3,474 & 464 \\
Slovenian    & Dev   & 0.67h & 400  & 80  \\
             & Test  & 0.99h & 358  & 36  \\
\midrule
             & Train & 1.34h & 1,265 &     \\
ParisStories & Dev   & 0.58h & 640  &  N/A  \\
             & Test  & 0.59h & 631  &     \\
\midrule
             & Train & 122.1h & 158,778 &     \\
Orféo & Dev   & 11.57h & 15,000  &  2000+  \\
             & Test  & 11.62h & 15,000  &     \\
\bottomrule
\end{tabular}%
}
\caption{Data statistics for the Naija-NSC treebank, Slovenian SST, ParisStories, and Orféo}
\label{tab:data_statistics}
\end{table}

\section{Related work}

Historically, speech parsing has mostly been addressed through parsing transcriptions, be the gold transcriptions or predicted transcriptions
\citep[e.g.\ ][]{bechet14_interspeech}.
Early work on speech parsing aimed at jointly detecting (and sometimes removing) disfluencies and parsing \citep{charniak-johnson-2001-edit,jorgensen-2007-effects,rasooli-tetreault-2013-joint,rasooli-tetreault-2014-non,honnibal-johnson-2014-joint,yoshikawa-etal-2016-joint,jamshid-lou-etal-2019-neural}.
The underlying motivation for removing the disfluencies is often to use the output tree as input to downstream NLP tools trained on written text and reduce the domain shift between speech and text.
Prior work also proposed to include audio features in text parsers, as an extra source of information to improve disambiguation \citep{kahn-etal-2005-effective,dreyer07_interspeech,pate-goldwater-2013-unsupervised,tran2021}.
In contrast, instead of relying on transcriptions, we address speech parsing directly from the speech signal and perform automatic speech recognition jointly, in line with \citet{pupier22_interspeech}.

Only a few publications have dealt with speech parsing without relying on transcriptions \citep{pupier22_interspeech,pupier-etal-2024-growing,kando24_interspeech}. They evaluate either on the French Orféo treebank \citep{benzitoun2016projet} for the first two, or Orféo and the English Switchboard treebank \citep{225858} for the last one.
This paper focuses instead on lower-resource languages: Naija and Slovenian. A second major novelty compared to these three publications is that we introduce a much simpler deep learning architecture.

Our work is also connected to \citet{lai2023audiovisualneuralsyntaxacquisition,10094575}, who investigate grammar inference from the speech signal and visual cues on a multimodal dataset of images with read captions.
However, the data they use consists of captions read aloud by speakers. 
Texts that are read aloud do not feature the phenomena that make speech recognition and analysis difficult (disfluencies, interruptions, overlapping speakers). Moreover, their syntax is similar to that of written texts.
As a result, there is a wide domain shift between spontaneous and read speech.
Our work focuses instead on corpora featuring realistic interactions.

\section{Datasets}
\label{sec:data}
We conduct a multilingual evaluation of Simplified W2T on three spoken treebanks from Universal Dependencies \citep{nivre-etal-2020-universal}, covering three languages. The multilingual evaluation here refers to training and evaluating the architecture separately on these three languages. We additionally use the French Orféo treebank  \citep{benzitoun2016projet} to assess the impact of scaling the training data on dependency parsing. The statistics of the datasets is presented in Table~\ref{tab:data_statistics}.

\begin{enumerate}[noitemsep, leftmargin=*]
  \item \textbf{Naija} - NaijaSynCor (Naija-SNC) treebank
  %v2.14
  \citep{caron2019surface}, released as part of Universal Dependencies \citep{nivre-etal-2020-universal}, was manually annotated using the Surface-Syntactic Universal Dependency \citep[SUD]{gerdes-etal-2018-sud} annotation scheme. Naija, officially named Nigerian Pidgin, is an English-based creole language, whose lexicon is influenced by Nigerian languages such as Yoruba, Hausa, and Igbo.  The treebank recordings span life stories, speeches, radio programs, free conversations, cooking recipes, comments on the current state
of affairs, etc.
  \item \textbf{Slovenian} - Slovenian SST \citep{dobrovoljc-nivre-2016-universal} was manually annotated using the Universal Dependencies (UD) annotation scheme. Its recordings cover multiple categories, including public informative and educational content, public entertainment, non-public non-private interactions, and non-public private conversations among friends or family members.
  \item  \textbf{UD French ParisStories\footnote{\url{https://universaldependencies.org/treebanks/
fr_parisstories/index.html}}} - The dataset was annotated following the SUD guidelines and automatically converted to UD. Its recordings consist of personal life stories.
  \item \textbf{French Orféo treebank}\footnote{Around 5\% of the Orféo treebank has been manually annotated (gold), the rest of the corpus is automatically annotated (good quality silver data)} - Orféo is the only publicly available spoken treebank large enough for scaling experiments. It has multiple sub-corpora, and the recordings span work meetings, interviews, public meetings, storytelling, narratives, and telephone conversations.
\end{enumerate}

We observed speech noise in the Slovenian-SST dataset which we filtered out, and enhanced speech-transcription misalignment.We provide further details in Appendix~\ref{sec:appendix_slovenian}. 

\begin{figure}[t]
  \begin{subfigure}{\columnwidth}
  \resizebox{\columnwidth}{!}{
  \includegraphics[]{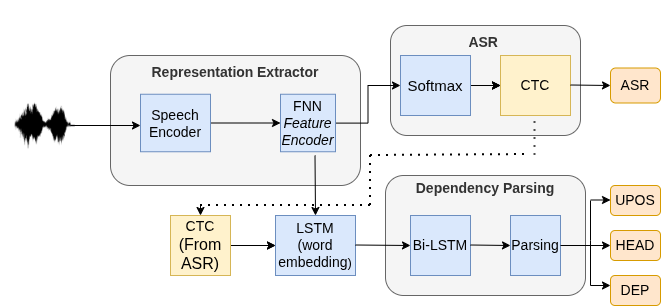}}
  \caption{Baseline Wav2tree \citep{pupier-etal-2024-growing}}
  \label{Baseline_wav2tree}
  \end{subfigure}
  \centering
  \begin{subfigure}{\columnwidth}
  \resizebox{\columnwidth}{!}{
  \includegraphics[]{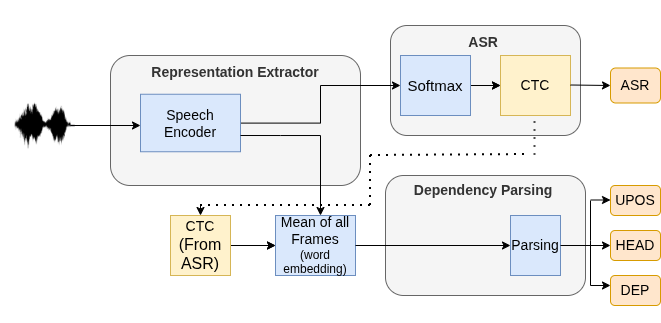}}
  \caption{Simplified Wav2tree (ours)}
  \label{Simplified_wav2tree}  
  \end{subfigure}
  \caption{Comparison of the original architecture \citep{pupier-etal-2024-growing} and our proposed architecture.}
\end{figure}

\section{Background: Wav2tree Architecture}
Wav2Tree \citep{pupier22_interspeech} is an end-to-end dependency parsing model whose only input is the raw speech signal. Its architecture is illustrated in Figure~\ref{Baseline_wav2tree}, and performs speech recognition and parsing jointly. 
The Wav2tree architecture is composed of three modules: (i) a speech encoder that computes a vector representation for each frame in the speech signal, (ii) an ASR module that outputs both a transcription and the frame boundaries of each predicted token, (iii) a parsing module that uses the word boundaries to compute audio word embeddings and run a standard parsing algorithm on the sequence of embeddings. These modules are connected by three intermediate NN units.

The audio signal $\mathbf s$ is first passed to a pre-trained speech encoder (Wav2Vec2, Eq.~\ref{eq:wav}) to obtain a sequence of frame vectors $\mathbf F^{(1)} \in \mathbb R^{\ell \times f}$, where ~$\ell$ is the length of the signal in number of frames and $f$ is the output size of Wav2Vec2 representations.
Then the obtained matrix is passed through the \nth{1} intermediate NN unit - Feedforward neural network (\textit{feature encoder}, Eq.~\ref{eq:fnn}).
\begin{align}
    \mathbf F^{(1)} &= \wavto ( \mathbf s ),    \label{eq:wav} \\
    \mathbf F^{(2)} &= \fnn (\mathbf F^{(1)}). \label{eq:fnn}
\end{align}
The speech recognition module applies a Connectionist temporal classification layer \citep[CTC]{graves2006connectionist} to $\mathbf F^{(2)}$. The granularity of the CTC prediction is at the character level: each frame is tagged with a character (or a special blank character representing a separation between two words). 

To obtain transcriptions from CTC predictions, Wav2tree uses a greedy decoding algorithm\footnote{\url{{https://github.com/Pupiera/ParseBrain/blob/master/parsebrain/speechbrain_custom/decoders/ctc.py}}} (Eq.~\ref{eq:ctc}). It is a modification of the greedy decode from the SpeechBrain library \citep{ravanelli2021speechbraingeneralpurposespeechtoolkit}, which outputs a single transcription. Along with the transcription $\hat t$, the decoding algorithm outputs a mapping $\hat a$ of the frames of Wav2Vec2 to the words, based on the character tagged on each frame. We provide an example of a mapping in Figure~\ref{fig:CTCexample} in Appendix. 
The \textit{predicted word boundaries} (start and end frames for each transcribed word) are extracted from this mapping. 

\begin{align}
    \hat t, \hat a &= \ctc ( \mathbf F^{(2)} ).    \label{eq:ctc}
    %\mathbf F^{(2)}_{\cdot, i_k:j_k} &= \alignment ( \mathbf F^{(2)}, \mathbf A) \label{eq:alignment}
\end{align}

The \nth{2} intermediate NN unit LSTM then uses these predicted word boundaries to compute the `audio' word embedding i.e.,\ fixed-sized vectors for each predicted word $w_k$ ($k \in \{1, \dots n\}$) with boundary indices $(i_k, j_k)$. This process is akin to using a character LSTM to compute textual word embedding based on their internal structure. \citep{ling-etal-2015-finding,plank-etal-2016-multilingual}. Instead of a sequence of characters, spoken words are a sequence of frames (the output of the \textit{feature encoder} module).
\begin{align}
    \mathbf w_{k} &= \lstm (\mathbf F^{(2)}_{\cdot, i_k:j_k}). \label{eq:audioword}
\end{align}
Next, Wav2tree computes contextualised audio word embedding by passing the sequence of embeddings $(\mathbf w_1, \dots, \mathbf w_n)$ through the \nth{3} intermediate NN unit - a bidirectional LSTM (Eq.\ \ref{eq:bilstm}):%
\begin{align}
\mathbf (\mathbf h_1, \dots, \mathbf h_n) &= \bilstm (\mathbf w_1, \dots, \mathbf w_n). \label{eq:bilstm}
\end{align}
The resulting word-level representations are passed to a generic dependency parsing algorithm, either a seq2label strategy \citep{strzyz-etal-2019-viable} or a biaffine architecture \citep{dozat2016deep}. 
The whole Wav2tree model is trained end-to-end with a multitask objective (ASR and parsing) to reduce the error propagation between the two tasks.

\section{Experiments}

The first set of experiments we conduct aims to answer the following research questions:
\begin{itemize}[leftmargin=*, noitemsep]
\item \textbf{RQ1}: What is the role of the intermediate neural network units in end-to-end parsing, and when do they add value?
\item \textbf{RQ2}: How is speech parsing accuracy affected by the size of the training data across two successive setups (i)  when the ASR model is fine-tuned jointly with the parsing model, and (ii) when a pre-fine-tuned ASR model is used to control for the effect of ASR accuracy.
\end{itemize}

The second set of experiments provides a comprehensive evaluation of speech parsing across the three spoken treebanks.

\subsection{RQ1: Role of Intermediate NN Units}

When the pre-trained model is fine-tuned jointly with the parsing model, its representations become aligned with the downstream parsing task. Building on this observation, we propose two hypotheses regarding the role of intermediate neural network units in the parsing architecture.
\begin{itemize}[leftmargin=*, noitemsep]
\item \textbf{H1}: When the pre-trained model is fine-tuned jointly with the parsing model,  the intermediate NN units do not add an incremental value.
\item \textbf{H2}: However, when the pre-trained model is frozen, the intermediate NN units help to reduce the \textit{representational gap} between the frozen pre-trained model and the parsing model.
\end{itemize}

\subsubsection{(H1) Simplified Architecture}
\label{sec:simplified}

In Wav2tree (Figure~\ref{Baseline_wav2tree}), the speech encoder is fine-tuned jointly with the CTC layer, the parsing model, and three intermediate NN units. We simplify the Wav2tree architecture by removing these intermediate NN units.

\begin{enumerate}[leftmargin=*, noitemsep]
  \item We removed the \textit{feature encoder} module. The representations from Wav2Vec2, learned via a self-supervised transformer-based architecture, largely encapsulate cross-segment dependencies without requiring further transformation. Therefore, in practice, we apply CTC directly to $\mathbf F^{(1)}$.
  
  \item For word-level representation, we directly use the representation from the speech model, by averaging all frames within each word. These representations are derived from a transformer-based encoder with self-attention, and thus integrate contextual information across the frames for a word. This suggests that the LSTM unit (Eq. \ref{eq:audioword}) is  redundant in this setting.
  
  \item We hypothesize that the bi-LSTM module is used to align the speech encoder representation with the parsing model. Since the architecture is fine-tuned end-to-end, i.e., the weights of the speech encoder are fine-tuned for the downstream parsing task, the bi-LSTM (Eq \ref{eq:bilstm}) is not required in this setting.

\end{enumerate}

The last two simplifications consist of feeding the sequence of vectors $(\mathbf F^{(1)}_{\text{mean}(i_1,j_1)}, \dots \mathbf F^{(1)}_{\text{mean}(i_n,j_n)})$ to the parser, effectively bypassing Eq. \ref{eq:audioword} and~\ref{eq:bilstm}.

The audio word embeddings $(\mathbf h_1, \dots, \mathbf h_n)$ in our architecture are computed by the three following equations (that replace Eq.\ \ref{eq:wav} to \ref{eq:bilstm}):
\begin{align}
    \mathbf F^{(1)} &= \wavto ( \mathbf s ),     \\
     \hat t, \hat a &= \ctc ( \mathbf F^{(1)} ),    \label{eq:ctcl} \\
     \mathbf (\mathbf h_1, \dots, \mathbf h_n) &= (\mathbf F^{(1)}_{\text{mean}(i_1,j_1)}, \dots \mathbf F^{(1)}_{ \text{mean}(i_n,j_n)}),
\end{align}
where $\hat a = [(i_1,j_1), \dots (i_n,j_n)]$ as previously mentioned, and $\mathbf F^{(1)}_{\text{mean}(i_k,j_k)}$ is the vector representation of the mean of all frames for a word $k$. 

In all, the output of speech representation from the speech encoder now directly serves as the input to the parsing model (see Figure~\ref{Simplified_wav2tree}). With this change, the parameter count is reduced by \textbf{12\%} in the simplified architecture, considering the hyperparameters used by \citet{pupier-etal-2024-growing}.

\subsubsection{(H2) When Do Intermediate NN Units Add Value?}
\label{sec:intermediate_nn}

We evaluate hypothesis H2 using a text parsing model. The text parsing architecture we use is the same as the Simplified architecture for speech parsing, except that the speech encoder is replaced by a pretrained text encoder. To study and isolate the impact of intermediate NN units, we feed gold transcriptions into the text parser. We do two experiments across all three spoken treebanks:

\begin{enumerate}[leftmargin=*, noitemsep]
\item \textit{Joint fine-tuning}: We first evaluate the text parsing performance when the architecture is fine-tuned end-to-end, with and without an intermediate NN unit (LSTM). \label{finetuningtextmodel}

\item \textit{Frozen pre-trained text model}: Next, we freeze the pre-trained text model (Figure~\ref{intermediate_nn}), and evaluate the text parsing performance with and without an intermediate NN unit.

\begin{figure}[t]
  \begin{subfigure}{\columnwidth}
  \resizebox{\columnwidth}{!}{
  \includegraphics[]{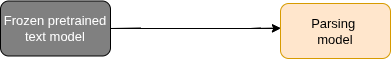}}
  \caption{No intermediate NN unit.}
  \label{no_nn}
  \end{subfigure}
%\end{figure}
%\begin{figure}
  \centering
  \begin{subfigure}{\columnwidth}
  \resizebox{\columnwidth}{!}{
  \includegraphics[]{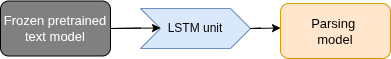}}
  \caption{With LSTM Unit}
  \label{with_nn}  
  \end{subfigure}
  \caption{Text parsing with frozen text encoder - with and without intermediate NN unit.}
  \label{intermediate_nn}
\end{figure}

\end{enumerate}

We note that it is not possible to conduct the same experiment for speech parsing, as the underlying speech model (Wav2Vec2) requires fine-tuning for the target language.

\subsection{RQ2: Scaling Training Data}
\label{sec:data_scaling}

We use the two largest spoken treebanks (Naija-SNC, 7h; Orféo, 122h) to assess the effect of the training set size.
The training data was partitioned into five subsets for this experiment: 20\%, 40\%, 60\%, 80\%, and 100\%.
With the simplified speech parsing architecture, we assess the amount of speech required to achieve an optimal performance. We conduct two-stage experiments for it:
\begin{enumerate}[noitemsep, leftmargin=*]
\item \textit{Baseline setup} \label{Scaling_baseline}:  We ran the Simplified W2T, without any modifications, across all data partitions.  
\item \textit{Frozen Speech encoder}: Since the input to parsing is the learned representation produced by the ASR model (speech encoder \& CTC tagging), ASR accuracy directly affects parsing performance. To isolate and precisely evaluate the effect of scaling training data on parsing, we use a pre-fine-tuned ASR system.
The pre-fine-tuned model is obtained from step~\ref{Scaling_baseline}, when trained with 100\% of the training data. Extending the finding from Section~\ref{sec:intermediate_nn}, we use an intermediate LSTM unit to align representation from the frozen ASR model to the parsing model.% and frozen when training the parsing module.

\end{enumerate}

\subsection{Dependency Parsing Evaluation}
\label{sec:compr_eval}
\paragraph{Middle Layers of Wav2Vec2} On the simplified architecture and for all three spoken treebanks, we assess the impact of the middle layers of the Wav2Vec2 on dependency parsing, i.e., using the representation from intermediate layers as the input to both ASR and parsing. This is motivated by the observations of \citet{shen23_interspeech} and \citet{dugonjic-etal-2024-lebenchmark}  who found through probing experiments that syntax is best encoded in the middle layers of the pre-trained speech models. 
\paragraph{Baseline experiments} On all three spoken treebanks, we checked the performance of text-based dependency parsing using a PIPELINE and a GOLD approach:
\begin{itemize}[noitemsep, leftmargin=*]
\item \textbf{PIPELINE}: The input to the text model is the predicted transcriptions from the pre-trained speech encoder of the Simplified architecture
\item \textbf{GOLD}: The inputs to the text parser are gold transcriptions. This experiment is meant as an upper bound (ideal case with perfect ASR).
\end{itemize}

\subsection{Segmentation Annotations in Naija}
\label{sec:segmentation}

The Naija dataset contains syntactic and prosodic segmentation annotations materialized by punctuation marks that are included as tokens in the dependency trees (see the tree in Figure~\ref{fig:tree} for an illustration). \citet{caron2019surface} found these markups useful for parsing, with gold transcriptions as the input.
We found this is not the case when parsing the audio signal (details in Appendix~\ref{sec:appendixpunctuations}).

\begin{table*}[t]
\begin{subtable}{\textwidth}
\resizebox{\textwidth}{!}{%
\begin{tabular}{lllrrrrl}
\toprule
\textbf{Model}   & \textbf{Wav2Vec2} & \textbf{Intermediate NN} & \multicolumn{1}{l}{\textbf{WER ↓}} & \multicolumn{1}{l}{\textbf{UPOS ↑}} & \multicolumn{1}{l}{\textbf{UAS ↑}} & \multicolumn{1}{l}{\textbf{LAS ↑}} & \textbf{Parameters} \\
 \midrule
Baseline W2T   & 24 layers                & yes                      & \textbf{31.4}                              & 77.8                               & \textbf{70.2 }                             & \textbf{62.7}                               & 315M$^w$+13M$^i$+4.5M$^p$    \\
Simplified W2T  & 24 layers              & no                       & 31.8                              & \textbf{79.1}                              & 69.1                               & 61.8                              & 315M$^w$+5M$^p$           \\
Simplified W2T & 21 layers               & no                       & 31.5                              & 78.5                               & 68.5                              & 61.3                              & 278M$^w$+5M$^p$           \\
\midrule
PIPELINE         & 24  layers             & no                       & 31.8                              & 77.9                               & 68.0                               & 61.2                              & 315M$^w$+5M$^p$+559M$^r$     \\
 \midrule
GOLD             & -               & no                       & 0.0                                & 96.9                               & 88.0                              & 83.9                              & 559M$^r$+5M$^p$          \\
 \bottomrule
\end{tabular}%
}
\caption{Evaluation on the Naija-NSC treebank.}
\label{tab:performance_naija}
\end{subtable}

\begin{subtable}{\textwidth}
\resizebox{\textwidth}{!}{%
\begin{tabular}{lllrrrrl}
\toprule
\multicolumn{1}{l}{\textbf{Model}} & \multicolumn{1}{l}{\textbf{Wav2Vec2}} & \multicolumn{1}{l}{\textbf{Intermediate NN}} & \multicolumn{1}{l}{\textbf{WER ↓}} & \multicolumn{1}{c}{\textbf{UPOS ↑}} & \multicolumn{1}{c}{\textbf{UAS ↑}} & \multicolumn{1}{l}{\textbf{LAS ↑}} & \multicolumn{1}{l}{\textbf{Parameters}} \\
\midrule
Baseline W2T                     & 24 layers                                    & yes                                          & 33.2                              & 73.0                               & 46.6                              & 37.7                               & 315M$^w$+13M$^i$+4.5M$^p$                        \\
Simplified W2T                    & 24 layers                                   & no                                           & \textbf{33.0}                              & \textbf{76.8}                               & 45.7                              & \underline{38.9}                              & 315M$^w$+5M$^p$                               \\
Simplified W2T                   & 18 layers                                    & no                                           & 34.3                              & 76.7                               & \textbf{47.4}                             & \textbf{40.7}                               & 240M$^w$+5M$^p$                               \\
\midrule
PIPELINE                           & 24 layers                                   & no                                           & 32.9                              & 80.2                               & 56.3                              & 50.9                              & 315M$^w$+5M$^p$+559M$^r$                         \\
\midrule
GOLD                               & -                                    & no                                           & 0.0                                & 97.8                               & 74.9                              & 71.3                              & 559M$^r$+5M$^p$   \\
\bottomrule
\end{tabular}%
}
\caption{Evaluation on the Slovenian treebank.}
\label{tab:performance_slovenian}
\end{subtable}

\begin{subtable}{\textwidth}
\resizebox{\textwidth}{!}{%
\begin{tabular}{lllrrrrl}
\toprule
\multicolumn{1}{l}{\textbf{Model}} & \multicolumn{1}{l}{\textbf{Speech\_large\_fr}} & \multicolumn{1}{l}{\textbf{Intermediate NN}} & \multicolumn{1}{l}{\textbf{WER ↓}} & \multicolumn{1}{l}{\textbf{UPOS ↑}} & \multicolumn{1}{l}{\textbf{UAS ↑}} & \multicolumn{1}{l}{\textbf{LAS ↑}} & \multicolumn{1}{l}{\textbf{Parameters}} \\
\midrule
Baseline W2T                     & 16 layers                                           & yes                                          & 38.2                              & 72.1                               & 58.4                              & 50.9                              & 313M$^g$+13M$^i$+4.5M$^p$                        \\
Simplified W2T                    & 16 layers                                          & no                                           & \textbf{37.6}                              & \textbf{73.6 }                              & 59.9                              & \underline{53.3}                              & 313M$^g$+5M$^p$                               \\
Simplified W2T                   & 11 layers                                          & no                                           & 38.1                              & 73.5                               & \textbf{60.7}                              & \textbf{54.2}                             & 250M$^g$+5M$^p$                               \\
\midrule
PIPELINE                           & 16 layers                                         & no                                           & 37.5                              & 71.7                               & 55.9                              & 48.7                              & 313M$^g$+5M$^p$+559M$^r$                         \\
\midrule
GOLD                               & -                                           & no                                           & 0.0                                & 96.1                               & 76.4                              & 71.9                              & 559M$^r$+5M$^p$                              \\
\bottomrule
\end{tabular}%
}
\caption{Evaluation on the ParisStories treebank.}
\label{tab:performance_parisstories}
\end{subtable}

\caption{Evaluation on \textbf{test} dataset with settings described in Section~\ref{sec:expe}. Parameter counts: $^w$Wav2vec, $^g$\texttt{Speech\_Large\_fr\_114K}, $^f$feedforward + LSTM, $^p$parsing module, $^r$Roberta module. W2T refers to Wav2tree. Best scores across end-to-end speech parsing highlighted in bold. Underlined refers to statistical significant results}
\label{tab:model_performance}
\end{table*}

\begin{table*}[]
\resizebox{\textwidth}{!}{%
\begin{tabular}{llrrrl}
\toprule

{\textbf{Text parsing architecture}} & {\textbf{Intermediate NN}} & {\textbf{LAS}} & {\textbf{LAS}} & {\textbf{LAS}} & {\textbf{\# of parameters}} \\
{} & {\textbf{(LSTM)}} & {\textbf{(Naija)}} & {\textbf{(Slovenian)}} & {\textbf{(ParisStories)}} & {(\textbf{fine-tuned)}} \\

\midrule
Joint fine-tuning                 & no                                           & \textbf{83.9}                                     & \textbf{71.3}                                       & \textbf{72.0}                                  & 559M$^r$+5M$^p$                                           \\
Joint fine-tuning + LSTM            & yes                                          & 81.7                                     & 66.1                                       & 69.8                                  & 559M$^r$+5M$^p$+16M$^i$                                        \\
\midrule
Frozen pre-trained text model                    & no                                           & 51.5                                     & 40.9                                       & 44.8                                  & 5M$^r$                                                 \\
Frozen pre-trained text model + LSTM               & yes                                          & \textbf{74.6}                                     & \textbf{55.4}                                       & \textbf{62.9}                                  & 5M$^p$+16M$^i$  \\
\bottomrule
\end{tabular}%
}
\caption{Comparative analysis for RQ1 - LAS on the test dataset with and without intermediate NN unit for all 3 treebanks with experiments described in Section~\ref{sec:intermediate_nn}. Best scores across sections highlighted in \textbf{bold}.}
\label{tab:text_parsing_lstm}
\end{table*}

\subsection{Training and Evaluation Details}
\label{sec:expe}

\paragraph{End-to-End speech parsing} 
We use \texttt{wav2vec2-xls-r-300m},\footnote{\url{https://huggingface.co/facebook/wav2vec2-xls-r-300m}} a multilingual Wav2Vec2 pre-trained acoustic model, as the speech encoder for both Naija-NSC treebank and Slovenian SST. It has Slovenian as medium-resource language (11k hours), and Yoruba \& Hausa (languages associated with Naija) as low-resource language (75 hours each) in its pre-training data. For UD French ParisStories, we use \texttt{Speech\_Large\_fr\_114K}, a data2vec \citep{pmlr-v162-baevski22a} pre-trained speech model trained on 114k hours of French speech  \citep{le-etal-2026-pantagruel}.

The parsing module is a biaffine graph-based parser \citep{dozat2016deep}. 
The baseline Wav2tree results are produced with the original Wav2tree implementation.

\paragraph{Text parsing} We use \texttt{xlm-roberta-large}\footnote{\url{https://huggingface.co/FacebookAI/xlm-roberta-large}} \citep{conneau2020unsupervisedcrosslingualrepresentationlearning} as text encoder for all three languages.  It has French as high-resource language (9.8 billion tokens), Slovenian as medium-resource language (1.7 billion tokens), and Hausa as low-resource language (56 million tokens) in its pre-training data.

We use the same hyperparameters across all architectures and datasets (see Table~\ref{tab:hyperparameters} in Appendix~\ref{sec:appendix} for the full list). We select the best checkpoint on the development set in each setting, and report the results on the test set. Our implementation uses the speechbrain library \citep{ravanelli2021speechbraingeneralpurposespeechtoolkit}.

\textbf{Evaluation metrics} We use standard evaluation measures: Word Error Rate (WER) for speech recognition, POS accuracy (POS), Unlabeled Attachment Score (UAS), and Labeled Attachment Score (LAS) for dependency parsing. The dependency parsing measures require an alignment between the ground truth and the predicted sequences, which was performed following the procedure described by \citet{pupier-etal-2024-growing}.

\section{Results}
\subsection{RQ1: Role of Intermediate NN Units}

\paragraph{(H1) Simplified architecture}

In Table \ref{tab:model_performance}, we show the comparison of baseline and simplified Wav2tree across all three spoken treebanks (\nth{1} and \nth{2} rows of each sub-table). 

After removing the three intermediate NN units, parsing accuracy (LAS) improved for Slovenian and French (Paris Stories). In absolute terms, LAS increased by 1.2\%, and 2.5\% on Slovenian and French. There is a slight drop for Naija (LAS decreased by 0.95\%). For Slovenian and French, the LAS improvement from the Baseline to the Simplified architecture is statistically significant (see Appendix~\ref{sec:results_appendix} for details). For all three datasets, ASR performance remains consistent across the Baseline and Simplified Wav2tree.

Based on this evaluation, we infer that the Simplified W2T, while being parameter-efficient (see Appendix~\ref{sec:computation-time} for the effect on training time), matches or improves parsing performance compared to Wav2tree. Our results support hypothesis H1 that intermediate NN units do not provide incremental value when the speech model is jointly fine-tuned with the parsing model.

\paragraph{(H2) When do intermediate NN units add value?}
\label{para:intermediate_nn_results }

In Table \ref{tab:text_parsing_lstm}, we report the text parsing performance (LAS score) on all three treebanks across the combinations of
(i) text parsing architectures with and without an intermediate NN unit (LSTM), and (ii) whether the pre-trained text model is frozen or fine-tuned jointly with the parsing model.

Across all three treebanks, when the pre-trained text model is fine-tuned jointly with the parsing model (\nth{1} and \nth{2} rows), LAS score is higher when there is no intermediate NN unit between the text encoder and the parsing model. This is consistent with our findings from the Simplified architecture. 

When we freeze the pre-trained text model and feed its output directly to the parsing model (\nth{3} row), the LAS score drops by at least 30\% across all three treebanks. We attribute this to the \textit{representational gap} between the frozen encoder and the parsing model. We then introduce an intermediate NN unit into this setup (\nth{4} row), and the LAS score increases by 15-20\% across all three treebanks, bringing the performance closer to the joint fine-tuning setting. This suggests that intermediate NN unit helps to reduce the \textit{representational gap} between the frozen pre-trained model and the parsing model.

We then measure the similarity between the text representation from the fine-tuned text model without any intermediate NN unit and (i) the representation from the frozen text model without any intermediate NN unit. The representation corresponds to the encoder output of the pre-trained model
(ii) the representation from the frozen text model with an intermediate NN unit. The representation corresponds to the output of the intermediate NN unit.

We use linear Centered Kernel Alignment \cite{kornblith2019similarity} as the similarity measure. Centered Kernel Alignment (CKA) is a widely used pairwise feature similarity for quantifying the similarity between two neural network representations. It measures the similarity between features from the two representations for each pair of examples using Hilbert-Schmidt Independence Criterion. 

The similarity is shown in Table \ref{tab:similaritymatruw}. Across all three treebanks, the similarity between the representations from the fine-tuned and the frozen text model is higher with an intermediate NN unit than without.

Based on these two evaluations, we answer RQ1 and support the hypothesis H2 that the intermediate NN unit helps to reduce the \textit{representational gap} between the frozen pre-trained model and the parsing model.

\subsection{RQ2: Scaling Training Data}

\begin{figure}[t]
  \begin{subfigure}{\columnwidth}
  \resizebox{\columnwidth}{!}{
  \includegraphics{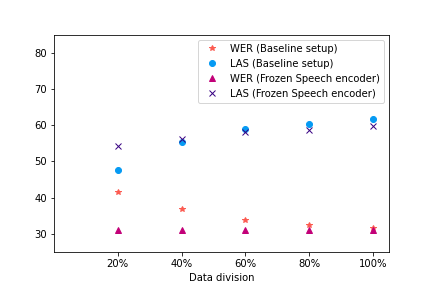}}
  \caption{Data Scaling - Naija }
  \label{ScalingNaija}
  \end{subfigure}
  \begin{subfigure}{\columnwidth}
  \resizebox{\columnwidth}{!}{
  \includegraphics[]{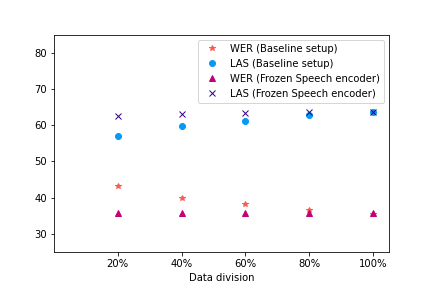}}
  \caption{Data Scaling - Orféo }
  \label{ScalingOrfeo}
  \end{subfigure}
  \caption{Scaling training data across 2 setups - End-to-End (Baseline) and Frozen ASR - as described in Section \ref{sec:data_scaling} }
  \label{fig:data_scaling}
\end{figure}

In Figure \ref{fig:data_scaling}, we report the performance of Naija and French (Orféo) across the two setups described in Section \ref{sec:data_scaling}. The WER is higher than usual for Orféo because of the nature of the
speech corpus (spontaneous discussions, and lower recording quality for some sub-corpora).

Across all treebanks, in the first experiment (\textit{Baseline setup}), the WER decreases and the LAS improves with a larger training set.
In the subsequent experiment (\textit{Frozen ASR}), where a pre-fine-tuned ASR is used across all data partitions, the LAS increases as the training data scales up, the magnitude of gains is reduced compared to the first experiment. Here, the LAS trend differs across the two spoken treebanks – the LAS remains relatively stable across all data partitions for Orféo, while it increases up to 80\% of the training data for Naija. We attribute this to the difference in the amount of the training data of the two treebanks. 

This analysis suggests that with a pre-fine-tuned ASR model, 
parsing accuracy gains diminish beyond 9-10 hours of training data with the Simplified W2T.

\subsection{Dependency Parsing Evaluation}
In Table \ref{tab:model_performance}, we report the evaluation for Wav2Vec2 intermediate layers, and the baseline results for all three spoken treebanks.

\paragraph{Intermediate layers}
The \nth{3} row in Table \ref{tab:model_performance} corresponds to the performance when the intermediate layer of the Wav2Vec2 model is used to produce the speech representation, as described in Section~\ref{sec:compr_eval}. 
We report results (on the test dataset) for the layer which has highest LAS score on the development dataset.  Across all three spoken treebanks, the LAS score remains stable or improved. This corresponds to a 12\%-24\% reduction in the parameters and 5-9\% speed up in training time. We conclude that intermediate layers of the speech encoder produce speech representations suitable for parsing.

\paragraph{Pipeline and GOLD}
The last two rows in Table~\ref{tab:model_performance} correspond to the Pipeline and the GOLD setups. Apart from the Slovenian treebank, the end-to-end Simplified architecture performs better than the Pipeline approach on parsing, which also has 2.75$\times$ more parameters than the Simplified architecture, due to the use of two pre-trained models: an acoustic model for ASR and a pretrained text model for parsing.

\paragraph{WER for Naija} We compare word error rate (WER) for tokens in Naija depending on whether they are in the Unix \texttt{words} English lexicon.\footnote{\url{https://www.nltk.org/book/ch02.html}} We find that WER is substantially higher for tokens not belonging to the English vocabulary, as displayed in Table~\ref{tab:In_distribution_Naija} in Appendix. 
This result is likely due to the limited size of Naija-related languages in its pretraining data, and illustrates the English bias of \texttt{wav2vec2-xls-r-300m}.

\begin{table}[]
\centering
\resizebox{\columnwidth}{!}{%
\begin{tabular}{llll}
\toprule
\textbf{Similarity between }   & \textbf{CKA}            & \textbf{CKA}             & \textbf{CKA}             \\
\textbf{fine-tuned encoder \&}               & \textbf{Naija}          & \textbf{Slovenian}       & \textbf{French}    \\
 \midrule
Frozen encoder \textbf{without}                            & \multicolumn{1}{r}{17\%} & \multicolumn{1}{r}{21\%} & \multicolumn{1}{r}{23\%} \\
  intermediate NN unit                                & {} & {} & {} \\
Frozen encoder \textbf{with}                   & \multicolumn{1}{r}{\textbf{44\%}} & \multicolumn{1}{r}{\textbf{32\%}} & \multicolumn{1}{r}{\textbf{34\%}} \\
intermediate NN unit                    & {} & {} & {} \\
\bottomrule
\end{tabular}%
}
\caption{Similarity between representations from fine-tuned and frozen text encoder described in Section~\ref{para:intermediate_nn_results }}
\label{tab:similaritymatruw}
\end{table}

\section{Conclusion and Further work}

We investigate the role of intermediate neural network units in the parsing architecture and find that they provide incremental value when the pre-trained model is frozen by reducing the \textit{representational gap} between the pre-trained model and parsing model. We further introduce an end-to-end parser for speech that is parameter-efficient while achieving comparable performance to the prior architecture. We evaluate the impact of scaling training data on the Simplified architecture, and find that gains in parsing accuracy diminish after the first 9-10h of training data.

We present baseline dependency parsing results for three spoken treebanks — Naija, Slovenian and French (ParisStories). We provide the first end-to-end speech parsing experiments for Naija and Slovenian.
We assess the impact of speech representation from intermediate layers of Wav2Vec2 as the input to the parsing architecture, and find that the parsing performance remains stable. Lastly, we release the code to reproduce filtered \texttt{UD\_Slovenian-SST} v2.16 treebank with improved speech-transcription alignment

The analysis of intermediate neural network units can be extended to other neural architectures beyond parsing. As the future work we plan to examine two findings from this paper: (i) why the pipeline performs better than the end-to-end speech parser on the Slovenian treebank, and (ii) why parsing performance remains stable when using representations from the intermediate layer of the speech encoder as the input.

\section*{Limitations}

For all our experiments, we limited our hyperparameter search to learning rate due to computational resource constraints. We do not evaluate the proposed architecture on the English Switchboard corpus \citep{225858} because the audio recordings are not publicly available. 

\section*{Acknowledgements}

We gratefully acknowledge the support of the French National Research Agency (ANR) through project SynPaX (grant ANR-23-CE23-0017-01).
We thank William Havard for feedback on an earlier version of this paper, as well as Rayan Ziane and Emmanuel Schang for fruitful discussions about this work.

\bibliography{custom}

@inproceedings{caron2019surface,
    title = "A Surface-Syntactic {UD} Treebank for {N}aija",
    author = "Caron, Bernard  and
      Courtin, Marine  and
      Gerdes, Kim  and
      Kahane, Sylvain",
    editor = "Candito, Marie  and
      Evang, Kilian  and
      Oepen, Stephan  and
      Seddah, Djam{\'e}",
    booktitle = "Proceedings of the 18th International Workshop on Treebanks and Linguistic Theories (TLT, SyntaxFest 2019)",
    month = aug,
    year = "2019",
    address = "Paris, France",
    publisher = "Association for Computational Linguistics",
    url = "https://aclanthology.org/W19-7803/",
    doi = "10.18653/v1/W19-7803",
    pages = "13--24"
}

@inproceedings{dozat2016deep,
  author       = {Timothy Dozat and
                  Christopher D. Manning},
  title        = {Deep Biaffine Attention for Neural Dependency Parsing},
  booktitle    = {5th International Conference on Learning Representations, {ICLR} 2017,
                  Toulon, France, April 24-26, 2017, Conference Track Proceedings},
  publisher    = {OpenReview.net},
  year         = {2017},
  url          = {https://openreview.net/forum?id=Hk95PK9le},
  bibsource    = {dblp computer science bibliography, https://dblp.org}
}

@inproceedings{strzyz-etal-2019-viable,
    title = "Viable Dependency Parsing as Sequence Labeling",
    author = "Strzyz, Michalina  and
      Vilares, David  and
      G{\'o}mez-Rodr{\'\i}guez, Carlos",
    booktitle = "NAACL 2019",
    year = "2019",
    url = "https://aclanthology.org/N19-1077",
}

@article{benzitoun2016projet,
  title={Le projet {ORF{\'E}O:} un corpus d’{\'e}tude pour le fran{\c{c}}ais contemporain},
  author={Benzitoun, Christophe and Debaisieux, Jeanne-Marie and Deulofeu, Henri-Jos{\'e}},
  journal={Corpus},
  number={15},
  year={2016},
  publisher={Bases, corpus et langage-UMR 6039}
}

@inproceedings{jorgensen-2007-effects,
    title = "The Effects of Disfluency Detection in Parsing Spoken Language",
    author = "J{\o}rgensen, Fredrik",
    editor = "Nivre, Joakim  and
      Kaalep, Heiki-Jaan  and
      Muischnek, Kadri  and
      Koit, Mare",
    booktitle = "Proceedings of the 16th Nordic Conference of Computational Linguistics ({NODALIDA} 2007)",
    month = may,
    year = "2007",
    address = "Tartu, Estonia",
    publisher = "University of Tartu, Estonia",
    url = "https://aclanthology.org/W07-2435",
    pages = "240--244",
}

@inproceedings{charniak-johnson-2001-edit,
    title = "Edit Detection and Parsing for Transcribed Speech",
    author = "Charniak, Eugene  and
      Johnson, Mark",
    booktitle = "Second Meeting of the North {A}merican Chapter of the Association for Computational Linguistics",
    year = "2001",
    url = "https://aclanthology.org/N01-1016",
}

@INPROCEEDINGS{225858,
  author={Godfrey, John J and Holliman, Edward C and McDaniel, Jane},
  booktitle={[Proceedings] ICASSP-92: 1992 IEEE International Conference on Acoustics, Speech, and Signal Processing}, 
  title={SWITCHBOARD: telephone speech corpus for research and development}, 
  year={1992},
  volume={1},
  number={},
  pages={517-520 vol.1},
  doi={10.1109/ICASSP.1992.225858}}

@inproceedings{rasooli-tetreault-2013-joint,
    title = "Joint Parsing and Disfluency Detection in Linear Time",
    author = "Rasooli, Mohammad Sadegh  and
      Tetreault, Joel",
    editor = "Yarowsky, David  and
      Baldwin, Timothy  and
      Korhonen, Anna  and
      Livescu, Karen  and
      Bethard, Steven",
    booktitle = "Proceedings of the 2013 Conference on Empirical Methods in Natural Language Processing",
    month = oct,
    year = "2013",
    address = "Seattle, Washington, USA",
    publisher = "Association for Computational Linguistics",
    url = "https://aclanthology.org/D13-1013",
    pages = "124--129",
}

@article{honnibal-johnson-2014-joint,
    title = "Joint Incremental Disfluency Detection and Dependency Parsing",
    author = "Honnibal, Matthew  and
      Johnson, Mark",
    editor = "Lin, Dekang  and
      Collins, Michael  and
      Lee, Lillian",
    journal = "Transactions of the Association for Computational Linguistics",
    volume = "2",
    year = "2014",
    address = "Cambridge, MA",
    publisher = "MIT Press",
    url = "https://aclanthology.org/Q14-1011",
    doi = "10.1162/tacl_a_00171",
    pages = "131--142",
}

@inproceedings{yoshikawa-etal-2016-joint,
    title = "Joint Transition-based Dependency Parsing and Disfluency Detection for Automatic Speech Recognition Texts",
    author = "Yoshikawa, Masashi  and
      Shindo, Hiroyuki  and
      Matsumoto, Yuji",
    editor = "Su, Jian  and
      Duh, Kevin  and
      Carreras, Xavier",
    booktitle = "Proceedings of the 2016 Conference on Empirical Methods in Natural Language Processing",
    month = nov,
    year = "2016",
    address = "Austin, Texas",
    publisher = "Association for Computational Linguistics",
    url = "https://aclanthology.org/D16-1109",
    doi = "10.18653/v1/D16-1109",
    pages = "1036--1041",
}

@article{pate-goldwater-2013-unsupervised,
    title = "Unsupervised Dependency Parsing with Acoustic Cues",
    author = "Pate, John K  and
      Goldwater, Sharon",
    editor = "Lin, Dekang  and
      Collins, Michael",
    journal = "Transactions of the Association for Computational Linguistics",
    volume = "1",
    year = "2013",
    address = "Cambridge, MA",
    publisher = "MIT Press",
    url = "https://aclanthology.org/Q13-1006",
    doi = "10.1162/tacl_a_00210",
    pages = "63--74",
}

@misc{lai2023audiovisualneuralsyntaxacquisition,
      title={Audio-Visual Neural Syntax Acquisition}, 
      author={Cheng-I Jeff Lai and Freda Shi and Puyuan Peng and Yoon Kim and Kevin Gimpel and Shiyu Chang and Yung-Sung Chuang and Saurabhchand Bhati and David Cox and David Harwath and Yang Zhang and Karen Livescu and James Glass},
      year={2023},
      eprint={2310.07654},
      archivePrefix={arXiv},
      primaryClass={cs.CL},
      url={https://arxiv.org/abs/2310.07654}, 
}

@INPROCEEDINGS{10094575,
  author={Tseng, Yuan and Lai, Cheng-I Jeff and Lee, Hung-Yi},
  booktitle={ICASSP 2023 - 2023 IEEE International Conference on Acoustics, Speech and Signal Processing (ICASSP)}, 
  title={Cascading and Direct Approaches to Unsupervised Constituency Parsing on Spoken Sentences}, 
  year={2023},
  volume={},
  number={},
  pages={1-5},
  doi={10.1109/ICASSP49357.2023.10094575}}

@inproceedings{kando24_interspeech,
  title     = {Textless Dependency Parsing by Labeled Sequence Prediction},
  author    = {Shunsuke Kando and Yusuke Miyao and Jason Naradowsky and Shinnosuke Takamichi},
  year      = {2024},
  booktitle = {Interspeech 2024},
  pages     = {1340--1344},
  doi       = {10.21437/Interspeech.2024-783},
  issn      = {2958-1796},
}

@inproceedings{kahn-etal-2005-effective,
    title = "Effective Use of Prosody in Parsing Conversational Speech",
    author = "Kahn, Jeremy G.  and
      Lease, Matthew  and
      Charniak, Eugene  and
      Johnson, Mark  and
      Ostendorf, Mari",
    editor = "Mooney, Raymond  and
      Brew, Chris  and
      Chien, Lee-Feng  and
      Kirchhoff, Katrin",
    booktitle = "Proceedings of Human Language Technology Conference and Conference on Empirical Methods in Natural Language Processing",
    month = oct,
    year = "2005",
    address = "Vancouver, British Columbia, Canada",
    publisher = "Association for Computational Linguistics",
    url = "https://aclanthology.org/H05-1030",
    pages = "233--240",
}

@inproceedings{bechet14_interspeech,
  title     = {Adapting dependency parsing to spontaneous speech for open domain spoken language understanding},
  author    = {Fr\'ed\'eric B\'echet and Alexis Nasr and Benoit Favre},
  year      = {2014},
  booktitle = {Interspeech 2014},
  pages     = {135--139},
  doi       = {10.21437/Interspeech.2014-39},
  issn      = {2958-1796},
}

@inproceedings{dreyer07_interspeech,
  title     = {Exploiting prosody for {PCFGs} with latent annotations},
  author    = {Markus Dreyer and Izhak Shafran},
  year      = {2007},
  booktitle = {Interspeech 2007},
  pages     = {450--453},
  doi       = {10.21437/Interspeech.2007-216},
  issn      = {2958-1796},
}

@inproceedings{pupier22_interspeech,
  title     = {End-to-End Dependency Parsing of Spoken {F}rench},
  author    = {Adrien Pupier and Maximin Coavoux and Benjamin Lecouteux and Jerome Goulian},
  year      = {2022},
  booktitle = {Interspeech 2022},
  pages     = {1816--1820},
  doi       = {10.21437/Interspeech.2022-381},
  issn      = {2958-1796},
}

@inproceedings{pupier-etal-2024-growing,
    title = "Growing Trees on Sounds: Assessing Strategies for End-to-End Dependency Parsing of Speech",
    author = "Pupier, Adrien  and
      Coavoux, Maximin  and
      Goulian, J{\'e}r{\^o}me  and
      Lecouteux, Benjamin",
    editor = "Ku, Lun-Wei  and
      Martins, Andre  and
      Srikumar, Vivek",
    booktitle = "Proceedings of the 62nd Annual Meeting of the Association for Computational Linguistics (Volume 2: Short Papers)",
    month = aug,
    year = "2024",
    address = "Bangkok, Thailand",
    publisher = "Association for Computational Linguistics",
    url = "https://aclanthology.org/2024.acl-short.22/",
    doi = "10.18653/v1/2024.acl-short.22",
    pages = "225--233"
}

@inproceedings{dugonjic-etal-2024-lebenchmark,
    title = "What Has {L}e{B}enchmark Learnt about {F}rench Syntax?",
    author = "Dugonji{\'c}, Zdravko  and
      Pupier, Adrien  and
      Lecouteux, Benjamin  and
      Coavoux, Maximin",
    editor = "Calzolari, Nicoletta  and
      Kan, Min-Yen  and
      Hoste, Veronique  and
      Lenci, Alessandro  and
      Sakti, Sakriani  and
      Xue, Nianwen",
    booktitle = "Proceedings of the 2024 Joint International Conference on Computational Linguistics, Language Resources and Evaluation (LREC-COLING 2024)",
    month = may,
    year = "2024",
    address = "Torino, Italia",
    publisher = "ELRA and ICCL",
    url = "https://aclanthology.org/2024.lrec-main.1521/",
    pages = "17493--17499"
}

@inproceedings{jamshid-lou-etal-2019-neural,
    title = "Neural Constituency Parsing of Speech Transcripts",
    author = "Jamshid Lou, Paria  and
      Wang, Yufei  and
      Johnson, Mark",
    editor = "Burstein, Jill  and
      Doran, Christy  and
      Solorio, Thamar",
    booktitle = "Proceedings of the 2019 Conference of the North {A}merican Chapter of the Association for Computational Linguistics: Human Language Technologies, Volume 1 (Long and Short Papers)",
    month = jun,
    year = "2019",
    address = "Minneapolis, Minnesota",
    publisher = "Association for Computational Linguistics",
    url = "https://aclanthology.org/N19-1282/",
    doi = "10.18653/v1/N19-1282",
    pages = "2756--2765"
}

@inproceedings{rasooli-tetreault-2014-non,
    title = "Non-Monotonic Parsing of Fluent Umm {I} mean Disfluent Sentences",
    author = "Rasooli, Mohammad Sadegh  and
      Tetreault, Joel",
    editor = "Wintner, Shuly  and
      Riezler, Stefan  and
      Goldwater, Sharon",
    booktitle = "Proceedings of the 14th Conference of the {E}uropean Chapter of the Association for Computational Linguistics, volume 2: Short Papers",
    month = apr,
    year = "2014",
    address = "Gothenburg, Sweden",
    publisher = "Association for Computational Linguistics",
    url = "https://aclanthology.org/E14-4010/",
    doi = "10.3115/v1/E14-4010",
    pages = "48--53"
}

@inproceedings{tran2021,
  author={Trang Tran and Mari Ostendorf},
  title={Assessing the Use of Prosody in Constituency Parsing of Imperfect Transcripts},
  year=2021,
  booktitle={Proc. Interspeech 2021},
  pages={2626--2630},
  doi={10.21437/Interspeech.2021-373}
}

@inproceedings{nivre-etal-2020-universal,
    title = "{U}niversal {D}ependencies v2: An Evergrowing Multilingual Treebank Collection",
    author = "Nivre, Joakim  and
      de Marneffe, Marie-Catherine  and
      Ginter, Filip  and
      Haji{\v{c}}, Jan  and
      Manning, Christopher D.  and
      Pyysalo, Sampo  and
      Schuster, Sebastian  and
      Tyers, Francis  and
      Zeman, Daniel",
    editor = "Calzolari, Nicoletta  and
      B{\'e}chet, Fr{\'e}d{\'e}ric  and
      Blache, Philippe  and
      Choukri, Khalid  and
      Cieri, Christopher  and
      Declerck, Thierry  and
      Goggi, Sara  and
      Isahara, Hitoshi  and
      Maegaard, Bente  and
      Mariani, Joseph  and
      Mazo, H{\'e}l{\`e}ne  and
      Moreno, Asuncion  and
      Odijk, Jan  and
      Piperidis, Stelios",
    booktitle = "Proceedings of the Twelfth Language Resources and Evaluation Conference",
    month = may,
    year = "2020",
    address = "Marseille, France",
    publisher = "European Language Resources Association",
    url = "https://aclanthology.org/2020.lrec-1.497/",
    pages = "4034--4043",
    language = "eng",
    ISBN = "979-10-95546-34-4"
}

@inproceedings{graves2006connectionist,
  title={Connectionist temporal classification: labelling unsegmented sequence data with recurrent neural networks},
  author={Graves, Alex and Fern{\'a}ndez, Santiago and Gomez, Faustino and Schmidhuber, J{\"u}rgen},
  booktitle={Proceedings of the 23rd international conference on Machine learning},
  pages={369--376},
  year={2006}
}

@inproceedings{ling-etal-2015-finding,
    title = "Finding Function in Form: Compositional Character Models for Open Vocabulary Word Representation",
    author = "Ling, Wang  and
      Dyer, Chris  and
      Black, Alan W  and
      Trancoso, Isabel  and
      Fermandez, Ram{\'o}n  and
      Amir, Silvio  and
      Marujo, Lu{\'i}s  and
      Lu{\'i}s, Tiago",
    editor = "M{\`a}rquez, Llu{\'i}s  and
      Callison-Burch, Chris  and
      Su, Jian",
    booktitle = "Proceedings of the 2015 Conference on Empirical Methods in Natural Language Processing",
    month = sep,
    year = "2015",
    address = "Lisbon, Portugal",
    publisher = "Association for Computational Linguistics",
    url = "https://aclanthology.org/D15-1176/",
    doi = "10.18653/v1/D15-1176",
    pages = "1520--1530"
}

@inproceedings{plank-etal-2016-multilingual,
    title = "Multilingual Part-of-Speech Tagging with Bidirectional Long Short-Term Memory Models and Auxiliary Loss",
    author = "Plank, Barbara  and
      S{\o}gaard, Anders  and
      Goldberg, Yoav",
    editor = "Erk, Katrin  and
      Smith, Noah A.",
    booktitle = "Proceedings of the 54th Annual Meeting of the Association for Computational Linguistics (Volume 2: Short Papers)",
    month = aug,
    year = "2016",
    address = "Berlin, Germany",
    publisher = "Association for Computational Linguistics",
    url = "https://aclanthology.org/P16-2067/",
    doi = "10.18653/v1/P16-2067",
    pages = "412--418"
}

@inproceedings{shen23_interspeech,
  title     = {Wave to Syntax: Probing spoken language models for syntax},
  author    = {Gaofei Shen and Afra Alishahi and Arianna Bisazza and Grzegorz Chrupała},
  year      = {2023},
  booktitle = {Interspeech 2023},
  pages     = {1259--1263},
  doi       = {10.21437/Interspeech.2023-679},
  issn      = {2958-1796},
}

@misc{ravanelli2021speechbraingeneralpurposespeechtoolkit,
      title={SpeechBrain: A General-Purpose Speech Toolkit}, 
      author={Mirco Ravanelli and Titouan Parcollet and Peter Plantinga and Aku Rouhe and Samuele Cornell and Loren Lugosch and Cem Subakan and Nauman Dawalatabad and Abdelwahab Heba and Jianyuan Zhong and Ju-Chieh Chou and Sung-Lin Yeh and Szu-Wei Fu and Chien-Feng Liao and Elena Rastorgueva and François Grondin and William Aris and Hwidong Na and Yan Gao and Renato De Mori and Yoshua Bengio},
      year={2021},
      eprint={2106.04624},
      archivePrefix={arXiv},
      primaryClass={eess.AS},
      url={https://arxiv.org/abs/2106.04624}, 
}

@inproceedings{chrupala-2023-putting,
    title = "Putting Natural in Natural Language Processing",
    author = "Chrupa{\l}a, Grzegorz",
    editor = "Rogers, Anna  and
      Boyd-Graber, Jordan  and
      Okazaki, Naoaki",
    booktitle = "Findings of the Association for Computational Linguistics: ACL 2023",
    month = jul,
    year = "2023",
    address = "Toronto, Canada",
    publisher = "Association for Computational Linguistics",
    url = "https://aclanthology.org/2023.findings-acl.495/",
    doi = "10.18653/v1/2023.findings-acl.495",
    pages = "7820--7827"
}

@inproceedings{dobrovoljc-nivre-2016-universal,
    title = "The {U}niversal {D}ependencies Treebank of Spoken {S}lovenian",
    author = "Dobrovoljc, Kaja  and
      Nivre, Joakim",
    editor = "Calzolari, Nicoletta  and
      Choukri, Khalid  and
      Declerck, Thierry  and
      Goggi, Sara  and
      Grobelnik, Marko  and
      Maegaard, Bente  and
      Mariani, Joseph  and
      Mazo, Helene  and
      Moreno, Asuncion  and
      Odijk, Jan  and
      Piperidis, Stelios",
    booktitle = "Proceedings of the Tenth International Conference on Language Resources and Evaluation ({LREC}'16)",
    month = may,
    year = "2016",
    address = "Portoro{\v{z}}, Slovenia",
    publisher = "European Language Resources Association (ELRA)",
    url = "https://aclanthology.org/L16-1248/",
    pages = "1566--1573"
}

@misc{pratap2023scalingspeechtechnology1000,
      title={Scaling Speech Technology to 1,000+ Languages}, 
      author={Vineel Pratap and Andros Tjandra and Bowen Shi and Paden Tomasello and Arun Babu and Sayani Kundu and Ali Elkahky and Zhaoheng Ni and Apoorv Vyas and Maryam Fazel-Zarandi and Alexei Baevski and Yossi Adi and Xiaohui Zhang and Wei-Ning Hsu and Alexis Conneau and Michael Auli},
      year={2023},
      eprint={2305.13516},
      archivePrefix={arXiv},
      primaryClass={cs.CL},
      url={https://arxiv.org/abs/2305.13516}, 
}

@article{demvsar2006statistical,
  title={Statistical comparisons of classifiers over multiple data sets},
  author={Dem{\v{s}}ar, Janez},
  journal={Journal of Machine learning research},
  volume={7},
  number={Jan},
  pages={1--30},
  year={2006}
}

@InProceedings{pmlr-v162-baevski22a,
  title = 	 {data2vec: A General Framework for Self-supervised Learning in Speech, Vision and Language},
  author =       {Baevski, Alexei and Hsu, Wei-Ning and Xu, Qiantong and Babu, Arun and Gu, Jiatao and Auli, Michael},
  booktitle = 	 {Proceedings of the 39th International Conference on Machine Learning},
  pages = 	 {1298--1312},
  year = 	 {2022},
  editor = 	 {Chaudhuri, Kamalika and Jegelka, Stefanie and Song, Le and Szepesvari, Csaba and Niu, Gang and Sabato, Sivan},
  volume = 	 {162},
  series = 	 {Proceedings of Machine Learning Research},
  month = 	 {17--23 Jul},
  publisher =    {PMLR},
  url = 	 {https://proceedings.mlr.press/v162/baevski22a.html}
}

@misc{conneau2020unsupervisedcrosslingualrepresentationlearning,
      title={Unsupervised Cross-lingual Representation Learning at Scale}, 
      author={Alexis Conneau and Kartikay Khandelwal and Naman Goyal and Vishrav Chaudhary and Guillaume Wenzek and Francisco Guzmán and Edouard Grave and Myle Ott and Luke Zettlemoyer and Veselin Stoyanov},
      year={2020},
      eprint={1911.02116},
      archivePrefix={arXiv},
      primaryClass={cs.CL},
      url={https://arxiv.org/abs/1911.02116}, 
}

@inproceedings{hull-dobrovoljc-2025-word,
    title = "Word Order Variation in Spoken and Written Corpora: A Cross-Linguistic Study of {SVO} and Alternative Orders",
    author = {H{\"u}ll, Nives  and
      Dobrovoljc, Kaja},
    editor = "Haji{\v{c}}ov{\'a}, Eva  and
      Kahane, Sylvain",
    booktitle = "Proceedings of the Eighth International Conference on Dependency Linguistics (Depling, SyntaxFest 2025)",
    month = aug,
    year = "2025",
    address = "Ljubljana, Slovenia",
    publisher = "Association for Computational Linguistics",
    url = "https://aclanthology.org/2025.depling-1.16/",
    pages = "150--155",
    ISBN = "979-8-89176-290-9"
}

@article{PeckBecker,
    url = {https://doi.org/10.1515/lingvan-2022-0156},
    title = {Syntactic pausing? Re-examining the associations},
    author = {Naomi Peck and Laura Becker},
    pages = {223--237},
    volume = {10},
    number = {1},
    journal = {Linguistics Vanguard},
    doi = {doi:10.1515/lingvan-2022-0156},
    year = {2024},
    lastchecked = {2026-03-17}
}

@inproceedings{kornblith2019similarity,
  title={Similarity of neural network representations revisited},
  author={Kornblith, Simon and Norouzi, Mohammad and Lee, Honglak and Hinton, Geoffrey},
  booktitle={International conference on machine learning},
  pages={3519--3529},
  year={2019},
  organization={PMlR}
}

@inproceedings{le-etal-2026-pantagruel,
  title = {Pantagruel: Unified Self-Supervised Encoders for French Text and Speech},
  author = {Le, Phuong-Hang and Pelloin, Valentin and Chatelain, Arnault and Bouziane, Maryem and Ghennai, Mohammed and Guan, Qianwen and Milintsevich, Kirill and Mdhaffar, Salima and Mannion, Aidan and Defauw, Nils and Gu, Shuyue and Audibert, Alexandre Daniel and Dinarelli, Marco and Estève, Yannick and Goeuriot, Lorraine and Lalande, Steffen and Hervé, Nicolas and Coavoux, Maximin and Portet, François and Ollion, Étienne and Candito, Marie and Peyrard, Maxime and Rossato, Solange and Lecouteux, Benjamin and Nardy, Aurélie and Sérasset, Gilles and Segonne, Vincent and Evain, Solène and Fabre, Diandra and Schwab, Didier},
  booktitle = {Proceedings of the Fifteenth Language Resources and Evaluation Conference (LREC 2026)},
  month = {May},
  year = {2026},
  pages = {10168--10191},
  address = {Palma, Mallorca, Spain},
  publisher = {European Language Resources Association (ELRA)},
  editor = {Piperidis, Stelios and Bel, Núria and van den Heuvel, Henk and Ide, Nancy and Krek, Simon and Toral, Antonio},
  doi = {10.63317/573q4exhmpgd}
}

@inproceedings{gerdes-etal-2018-sud,
    title = "{SUD} or Surface-Syntactic {U}niversal {D}ependencies: An annotation scheme near-isomorphic to {UD}",
    author = "Gerdes, Kim  and
      Guillaume, Bruno  and
      Kahane, Sylvain  and
      Perrier, Guy",
    editor = "de Marneffe, Marie-Catherine  and
      Lynn, Teresa  and
      Schuster, Sebastian",
    booktitle = "Proceedings of the Second Workshop on Universal Dependencies ({UDW} 2018)",
    month = nov,
    year = "2018",
    address = "Brussels, Belgium",
    publisher = "Association for Computational Linguistics",
    url = "https://aclanthology.org/W18-6008/",
    doi = "10.18653/v1/W18-6008",
    pages = "66--74"
}

@article{baevski2020wav2vec,
  title={wav2vec 2.0: A framework for self-supervised learning of speech representations},
  author={Baevski, Alexei and Zhou, Yuhao and Mohamed, Abdelrahman and Auli, Michael},
  journal={Advances in neural information processing systems},
  volume={33},
  pages={12449--12460},
  year={2020}
}

\appendix

\section{Hyperparameters}
\label{sec:appendix}

We provide the full list of hyperparameters for our experiments in Table~\ref{tab:hyperparameters}.

\begin{table*}[]
\centering
\resizebox{\linewidth}{!}{%
\begin{tabular}{lrr}
\toprule
\textbf{Architecture} &
  \multicolumn{1}{r}{\textbf{Baseline Wav2tree  \citep{pupier-etal-2024-growing}}} &
  \multicolumn{1}{r}{\textbf{Simplified Wav2tree (ours)}} \\
 \midrule
Epoch                      & 80        & 80       \\
Batch size                 & 4         & 4        \\
\midrule
\textbf{Tuning parameters} &           &          \\
Learning rate (ASR)              & 0.00005    & 0.00005   \\
Learning rate (Remaining)            & 0.0003    & 0.0003   \\
Optimizer                  & AdamW  & AdamW \\
Model name &
  \begin{tabular}[c]{@{}r@{}}wav2vec2-xls-r-300m (Naija \& Slovenian)\\ PantagrueLLM/Speech\_Large\_fr\_114K (ParisStories)\end{tabular} &
  \begin{tabular}[c]{@{}r@{}}wav2vec2-xls-r-300m (Naija \& Slovenian)\\ PantagrueLLM/Speech\_Large\_fr\_114K (ParisStories)\end{tabular} \\
\midrule
\textbf{Encoder}           & Present   & N/A      \\
Encoder layer              & 3         &          \\
Dropout                    & 0.15      &          \\
Encoder dim                & 1024      &          \\
Activation                 & LeakyReLU &          \\
\midrule
\textbf{Fusion LSTM}       & Present   & N/A      \\
Layer                      & 2         &          \\
Dim                        & 500       &          \\
Bidirectional              & FALSE     &          \\
Bias                       & TRUE      &          \\
\midrule
\textbf{LSTM parser}       & Present   & N/A      \\
Layer                      & 3         &          \\
Dim                        & 256       &          \\
Bidirectional              & TRUE      &          \\
\midrule
\textbf{Arc MLP}           &           &          \\
Dim                        & 256       & 256      \\
Layer                      & 1         & 1        \\
Linear head dim            & 256       & 256      \\
\midrule
\textbf{Label MLP}         &           &          \\
Dim                        & 256       & 256      \\
Layer                      & 1         & 1        \\
Head dim                   & 256       & 256      \\
\midrule
\textbf{POS MLP}           &           &          \\
Dim                        & 256       & 256      \\
Linear head dim &
  Dependent on the data &
  Dependent on the data \\
 \bottomrule
\end{tabular}%
}
\caption{Hyperparameters across Wav2tree architecture \citep{pupier-etal-2024-growing} and Simplified Wav2tree (ours). }
\label{tab:hyperparameters}
\end{table*}

\begin{figure*}
    
    \resizebox{\textwidth}{!}{
        \includegraphics{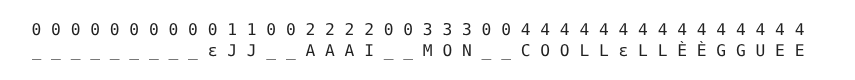}
    }
    \caption{Predicted word boundary and the corresponding uncollapsed sequence of the predicted transcription (in French) "J AI MON COLLÈGUE" (\textsc{en:} I have my colleague). Symbols $\epsilon$  and \_ represents respectively the blank and space labels. Indices \{1,2,3,4\} refer to the word position in the output sequence.}
    \label{fig:CTCexample}
\end{figure*}

\begin{table}[]
\resizebox{\columnwidth}{!}{%
\begin{tabular}{llrrrr}
\toprule
\textbf{Index} & \multicolumn{1}{c}{\textbf{Punctuation presence}} & \multicolumn{1}{c}{\textbf{WER ↓}} & \multicolumn{1}{c}{\textbf{UPOS ↑}} & \multicolumn{1}{c}{\textbf{UAS ↑}} & \multicolumn{1}{c}{\textbf{LAS ↑}} \\
\midrule
1 & Train \& Eval  data                                   & 34.4                               & 76.5                                & 62.8                               & 57.6                               \\
2 & Training data                                              & 32                                 & 78.5                                & 68.8                               & \textbf{61.8}                               \\
3 & None                                               & \textbf{31.8}                             & \textbf{79.1}                             & \textbf{69.1}                               & \textbf{61.8}     \\ 
\bottomrule
\end{tabular}%
}
\caption{Simplified W2T results across different variations of Naija dataset. Best scores highlighted in \textbf{bold}.}
\label{tab:punctuation_results}  
\end{table}

\begin{figure}[t]
  \begin{subfigure}{\columnwidth}
  \resizebox{\columnwidth}{!}{
  \includegraphics[]{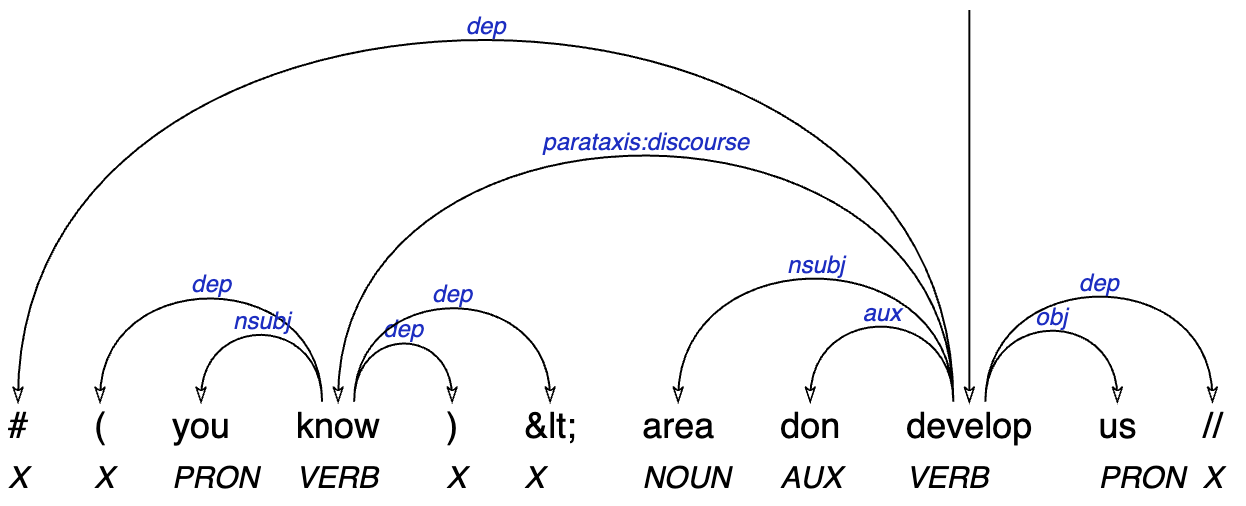}}
  \caption{Gold annotation}
  \label{gold_annotation}
  \end{subfigure}
%\end{figure}
%\begin{figure}
  \centering
  \begin{subfigure}{\columnwidth}
  \resizebox{\columnwidth}{!}{
  \includegraphics[]{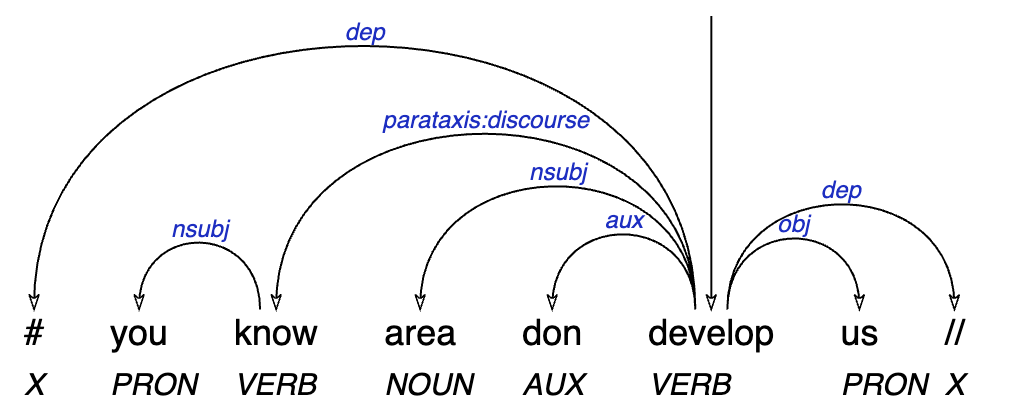}}
  \caption{Predicted annotation}
  \label{predicted_annotation}  
  \end{subfigure}
  \caption{Gold and predicted transcription and trees for a Naija utterance (sent\textunderscore id: WAZL\textunderscore 15\textunderscore MC\textunderscore Abi-MG\textunderscore \textunderscore 180). The predicted annotation has all UAS, UPOS and LAS as 72.7. The scores are low because not all markups are predicted by ASR - `(', `)' and `$<$'(marked as `\&lt') are not predicted by ASR.}
\label{segmentation_example}  
\end{figure}

\begin{table}[h]
\centering
\resizebox{\columnwidth}{!}{%
\begin{tabular}{lrr}
\toprule
\multicolumn{1}{l}{\textbf{English Vocab}} & \multicolumn{1}{l}{\textbf{Number of words}} & \multicolumn{1}{l}{\textbf{WER}} \\
\midrule
1                                                         & 8,991                                               & 24.5\%                           \\
0                                                         & 1,269                                               & 52.5\%       \\  
\bottomrule
\end{tabular}%
}
\caption{WER for tokens in Naija-NSC treebank.}
\label{tab:In_distribution_Naija}
\end{table}

\section{Punctuation in Naija}
\label{sec:appendixpunctuations}
The main punctuation marks in Naija dataset are:\footnote{We refer the reader to \citet{caron2019surface} for a more thorough description and examples.}
\begin{itemize}[noitemsep,  leftmargin=*]
    \item \texttt{//} delimits illocutionary units (final token of each dependency tree)
    \item a bracketing with \texttt{\{ \}} indicates a sequence of elements with the same syntactic function; elements in the sequence are delimited by a separator that depends on the type of the sequence. For example, in Figure~\ref{fig:tree}, a disfluency with a false start is represented with the pattern \texttt{\{ reparandum || repair \} }. Possible separators are \texttt{||} for reformulations and disfluencies, \texttt{|c} for coordination, \texttt{|r} for syntactic reduplication and \texttt{|a} for apposition.
    \item \texttt{\#} denotes a short pause.
    \item \texttt{<} and \texttt{>} separate the central part of the utterance (nucleus) from the peripheral elements \citep[ad-nuclei, see][and Figure~\ref{gold_annotation}]{caron2019surface}.
\end{itemize}

We created three setups to investigate the impact of the punctuation marks for speech parsing.
\begin{enumerate}[noitemsep,  leftmargin=*]
    \item Both training and evaluation datasets have markups. \label{with}
    \item Only training dataset has markups. During evaluation, markups are removed from the prediction before computing ASR and parsing measures \label{partial}
    \item Neither training nor evaluation datasets have markups \label{nopunctuations}
\end{enumerate}

We identify punctuation marks as characters that do not contain alphabetic characters or numbers. 
We first assess the utility of punctuation marks in the evaluation dataset (comparing the first 2 setups). This evaluation is aligned with the the comparison setting of \citet{caron2019surface}, who evaluated the parsing performance on treebank with and without markups. Table \ref{tab:punctuation_results} shows this comparison (\nth{1} and \nth{2} rows). The LAS score drops by 4.1\% (in absolute terms) if there are punctuation marks in the evaluation dataset. 

We dig deeper into why the presence of segmentation annotations in the transcription have a negative impact on parsing. We find a correlation between the ASR punctuation recognition errors and parsing accuracy. Specifically, as the number of markups increases in the transcript, not all these markups are predicted by ASR, which impacts the performance of both ASR and parsing. Table~\ref{tab:With_without_punctuations} displays the performance of evaluation data with and without markups, stratified by the number of markups present in the ground truth transcript. As the number of markups in the transcript increases, ASR predicts fewer markups than present in the ground truth, and this degradation also affects LAS, as evidenced by the increasing widening gap between LAS with and without markups. Figure~\ref{segmentation_example} shows a similar illustration by an example from the Naija treebank. We hypothesize that recognition errors for segmentation symbols arise because these markups do not correspond to identifiable sounds in the audio signal.

\citet{caron2019surface} found the punctuation marks useful for parsing, which can be because they used gold transcriptions, an ideal case when parsing transcriptions.

We then assess if having punctuation in the training dataset helps parser to do a better demarcation (comparing the last 2 setups). The \nth{2} and \nth{3} rows in Table \ref{tab:punctuation_results} show this comparison. The parsing performance remains same or improved with no punctuation marks in the training dataset, and thus we consider the Naija dataset without punctuation for all our evaluations.

\begin{table*}[]
\centering
\resizebox{\textwidth}{!}{%
\begin{tabular}{lrrrrrr}
\toprule
\multicolumn{1}{l}{\textbf{\begin{tabular}[c]{@{}l@{}}\# of \\ markups\end{tabular}}} &
  \multicolumn{1}{l}{\textbf{Trees}} &
  \multicolumn{1}{l}{\textbf{\begin{tabular}[c]{@{}l@{}}Avg markups \\ in ground\_truth\end{tabular}}} &
  \multicolumn{1}{l}{\textbf{\begin{tabular}[c]{@{}l@{}}Avg markups \\ in prediction\end{tabular}}} &
  \multicolumn{1}{l}{\textbf{\begin{tabular}[c]{@{}l@{}}LAS with \\ markups\end{tabular}}} &
  \multicolumn{1}{l}{\textbf{\begin{tabular}[c]{@{}l@{}}LAS without \\ markups\end{tabular}}} &
  \multicolumn{1}{l}{\textbf{\begin{tabular}[c]{@{}l@{}}LAS \\ decrease\end{tabular}}} \\
  \midrule
{[}0, 2{]}        & 423 & 1.5  & 1.9  & 66.0 & 66.0 & 0.0   \\
{[}3, 4{]}        & 232 & 3.4  & 3.5  & 63.0 & 64.5 & -1.4  \\
{[}5, 6{]}        & 137 & 5.4  & 4.8  & 57.1 & 61.1 & -3.9  \\
{[}7, 12{]}       & 139 & 8.7  & 7.1  & 54.2 & 60.3 & -6.1  \\
\textgreater{}=13 & 41  & 16.5 & 12.6 & 44.0 & 54.0 & -10.0 \\
\bottomrule
\end{tabular}%
}
\caption{Parsing results comparison with and without syntax-prosody segmentation annotations.}
\label{tab:With_without_punctuations}
\end{table*}

\section{Slovenian dataset}
\label{sec:appendix_slovenian}
Learning from our findings on the Naija punctuation marks, we have removed the punctuations from the Slovenian dataset for our analysis. Using the current version of the UD Slovenian dataset, we observed a low parsing performance (30\% LAS). After investigation, we identified two types of noise, which we filtered out. Additionally, we filtered out noisy recordings.
\begin{enumerate}
 \item \textbf{Speech misalignments noise}: Multiple transcripts are sometimes associated with a single speech recording without start or end timestamps.
 
 \item \textbf{Transcription artefacts noise}: The transcripts include placeholder tokens for personal names (eg: \textit{[name:personal], [name:surname]}), which mislead the ASR system and consequently degrade parsing performance. The corresponding audio has \textit{beeps} for it.
  
\end{enumerate}

For unique recordings, the speech–transcript alignment is not fully accurate, and we used forced alignment \citep{pratap2023scalingspeechtechnology1000} for it.

\section{Additional Results}
\label{sec:results_appendix}
\subsection{Statistical Significance}

Along with the parsing measures comparison, we perform significance testing on the test dataset of all 3 treebanks. We use the non-parametric Wilcoxon signed rank test \citep{demvsar2006statistical} with the parsing measure LAS score as the measurement. We first check for the \textbf{null} hypothesis that there is no statistically significant difference between the performance of the simplified and the baseline architectures. The \nth{1} row in Table \ref{tab:significance_testing} has the p-value for this null hypothesis. For Slovenian and ParisStories, we reject the null hypothesis at a confidence level of 5\%, and conclude that there is a difference in the performance of 2 architectures. For Naija, this null hypothesis cannot be rejected.

For Slovenian and ParisStories, following that simplified and baseline architectures are statistically different, we checked for an \textbf{alternative} hypothesis that the LAS score of ‘simplified’ architecture is greater than the ‘baseline’ architecture (\nth{2} row in Table \ref{tab:significance_testing}). Here, the null hypothesis is that the difference in LAS score between simplified and baseline is less than a distribution symmetrical about zero. For these 2 treebanks, this null hypothesis can be rejected at a confidence level of 5\% , in favor of the alternative that the difference in LAS score between simplified and baseline architecture is greater than the distribution symmetric about zero.

\begin{table}[h]
\centering
\resizebox{\columnwidth}{!}{%
\begin{tabular}{llrrr}
\toprule
{\textbf{Index}} &{\textbf{Hypothesis}} & {\textbf{p value}} & {\textbf{p value}}  & {\textbf{p value}} \\
& & \textbf{Naija} & \textbf{Slovenian} & \textbf{ParisStories} \\
\midrule
1                                                         & Null                                               & 21.6\% & 0.2\% & 3.6\%                           \\
2                                                         & Alternative   & -                                          & 0.1\% & 1.8\%        \\  
\bottomrule
\end{tabular}%
}
\caption{Statistical significance testing between Simplified and Baseline architecture}
\label{tab:significance_testing}
\end{table}

\section{Computational Time}
\label{sec:computation-time}
All experiments were conducted on an NVIDIA H100 NVL GPU (96GB memory) with 64 cores. The total computational budget for this paper (including preliminary experiments) was 800 GPU hours.

The computation times for training and inference with the baseline and simplified Wav2tree on the Naija dataset are presented in  Table~\ref{tab:Computational_time}.
The simplified architecture (ours) demonstrates a significant reduction in training time compared to the baseline architecture \citep[Wav2tree]{pupier-etal-2024-growing}. On the Naija corpus, and under identical hyperparameter settings, the simplified version requires only 60\% of the training computational time needed by the baseline architecture. The inference time is also reduced by 20\% with the simplified architecture.

Furthermore, with the simplified architecture, using intermediate layers of Wav2Vec2 (21 layers) results in an 5\% reduction in training time compared to using the full set of layers.

\begin{table}[h]
\centering
\resizebox{\columnwidth}{!}{%
\begin{tabular}{llrrr}
\toprule
{\textbf{Index}} &{\textbf{Hypothesis}} & {\textbf{p value}} & {\textbf{p value}}  & {\textbf{p value}} \\
& & \textbf{Naija} & \textbf{Slovenian} & \textbf{ParisStories} \\
\midrule
1                                                         & Null                                               & 21.6\% & 0.2\% & 3.6\%                           \\
2                                                         & Alternative   & -                                          & 0.1\% & 1.8\%        \\  
\bottomrule
\end{tabular}%
}
\caption{Statistical significance testing between Simplified and Baseline architecture}
\label{tab:significance_testing}
\end{table}

\begin{table}[]
\centering
\resizebox{\columnwidth}{!}{%
\begin{tabular}{llcc}
\toprule
\multicolumn{1}{l}{\textbf{Wav2tree}} &
  \multicolumn{1}{c}{\textbf{Wav2Vec2}} &
  \multicolumn{1}{c}{\textbf{Training}} &
  \multicolumn{1}{c}{\textbf{Inference}} \\
%& & & \textbf{Time} & \textbf{Time} \\
  \midrule
Baseline   & 24 layers & 676min & 25s \\
Simplified & 24 layers & 403min & 20s \\
Simplified & 21 layers & 381min & 20s \\
\bottomrule
\end{tabular}%
}
\caption{Computation time on the Naija corpus.}
\label{tab:Computational_time}
\end{table}

\end{document}